\documentclass{article}
\usepackage[accepted,nohyperref]{icml2019}
\usepackage{microtype}
\usepackage{graphicx}
\usepackage{subfigure}
\usepackage{booktabs} 

\usepackage{amsthm}
\usepackage{multicol}
\usepackage{multirow}
\usepackage[table]{xcolor}
\usepackage{booktabs}

\usepackage{url}
\usepackage{lineno}
\usepackage[onehalfspacing]{setspace}

\usepackage[colorinlistoftodos,prependcaption,textsize=tiny]{todonotes}

\usepackage[toc,page]{appendix}

\usepackage{color,soul}
\soulregister\cite7
\soulregister\citep7
\soulregister\ref7
\usepackage{natbib}
\usepackage{graphicx}

\def\BN{Batch Normalization}
\def\BNp{Batch Normalization\ }
\def\LNp{Local Normalization\ }

\icmltitlerunning{LocalNorm: Robust Image Classification through Dynamically Regularized Normalization}

\begin{document}

\twocolumn[
\icmltitle{LocalNorm: Robust Image Classification through Dynamically Regularized Normalization}



\icmlsetsymbol{equal}{*}

\begin{icmlauthorlist}
\icmlauthor{Bojian Yin}{cwi}
\icmlauthor{Siebren Schaafsma}{holst}
\icmlauthor{Henk Corporaal}{tue}
\icmlauthor{H. Steven Scholte}{uvaa}
\icmlauthor{Sander M. Bohte}{cwi,uvab,rug}
\end{icmlauthorlist}

\icmlaffiliation{cwi}{Machine Learning group, Centrum Wiskunde \& Informatica (CWI), Amsterdam, The Netherlands}
\icmlaffiliation{uvaa}{Dept of Psychology, University of Amsterdam, Amsterdam, The Netherlands}
\icmlaffiliation{uvab}{Dept of Biology, University of Amsterdam, Amsterdam, The Netherlands}
\icmlaffiliation{tue}{Dept of Electrical Engineering, Technical University Eindhoven, Eindhoven, The Netherlands}
\icmlaffiliation{holst}{Holst Centre / IMEC, Eindhoven, The Netherlands}
\icmlaffiliation{rug}{Dept AI, Rijksuniversiteit Groningen, Groningen, The Netherlands}

\icmlcorrespondingauthor{Bojian Yin}{Bojian.Yin@cwi.nl}
\icmlcorrespondingauthor{Sander Bohte}{S.M.Bohte@cwi.nl}


\vskip 0.3in
]



\printAffiliationsAndNotice{}  

\begin{abstract}
While modern convolutional neural networks achieve outstanding accuracy on many image classification tasks, they are, compared to humans, much more sensitive to image degradation. Here, we describe a variant of Batch Normalization, LocalNorm, that regularizes the normalization layer while dynamically adapting to the local image intensity and contrast at test-time. We show that the resulting networks are much more resistant to noise-induced image degradation, improving accuracy by up to three times, while achieving the same or better accuracy on non-degraded classical benchmarks. We also show that LocalNorm is more robust to image distortions in general, as measured on the CIFAR10-c dataset. In computational terms, LocalNorm can be applied to single images at test-time, adds negligible training cost and little or no cost at inference time, and can be applied to already-trained networks in a straightforward manner. 
\end{abstract}


\section{Introduction}
Methods that reduce internal covariate shift via learned rescaling and recentering neural activation, like \BNp \cite{ioffe2015batch}, have been an essential ingredient for successfully training deep neural networks (DNNs). In \BN, neural activation values are rescaled with trainable parameters, where summary neural activity is typically computed as mean and standard deviation over a batch of inputs. Such compact batch statistics however are sensitive to the input distribution, resulting in errors when novel images are outside this distribution, for example when faced with different and unseen lighting or noise conditions. Then, and unlike the human visual system, modern DNNs perform and generalize poorly 
\cite{geirhos2018generalisation}. 

\begin{figure}[b!]
    \centering
    \vspace*{-7mm}
    \includegraphics[width=\columnwidth,height=3.6cm]{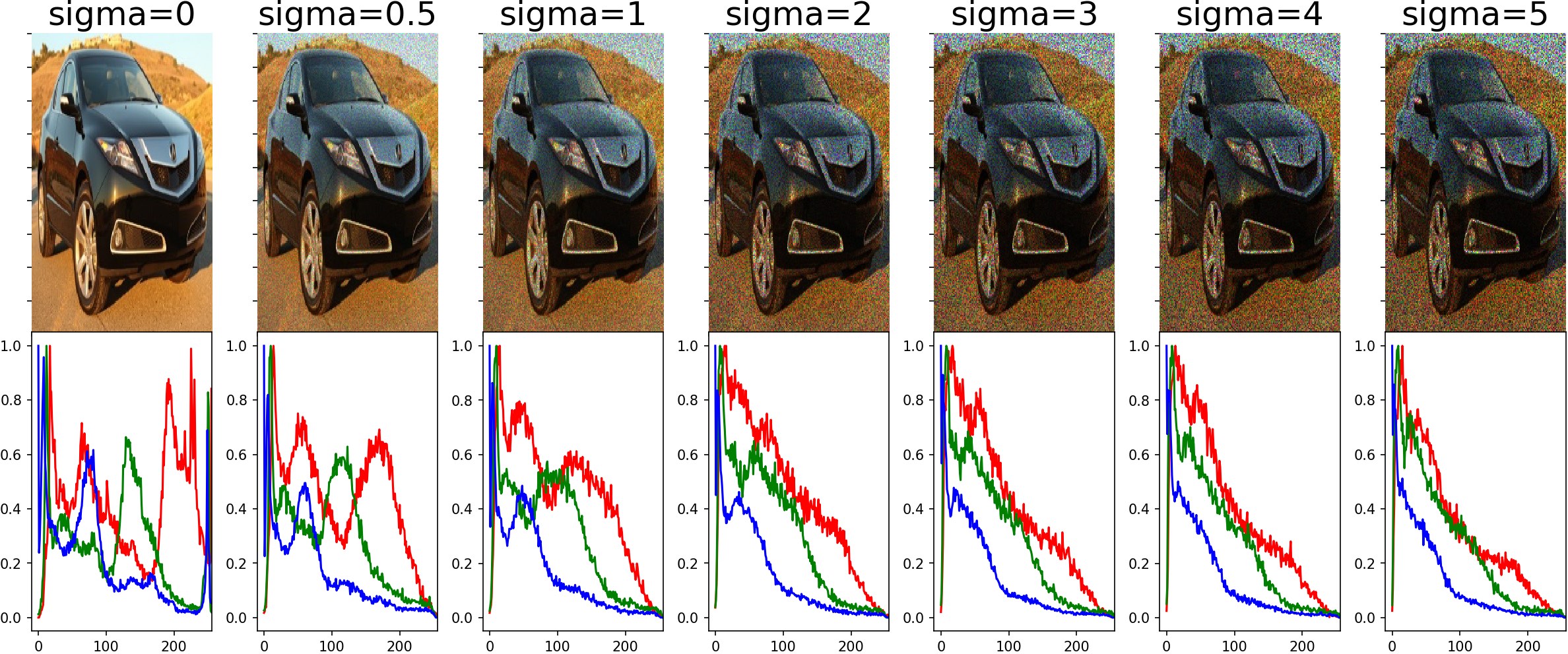}
    \vspace*{-7mm}
    \caption{RGB-Histogram for increasing additive Gaussian noise}
    \label{fig:agn-rgb}
\end{figure}

While the original \BNp computed statistics across the activity in a single feature map (or {\em channel}) \cite{ioffe2015batch}, trainable normalizations have been proposed along a number of dimensions of deep neural network layers, including {Layer Normalization}, \cite{ba2016layer}, Group Normalization \cite{wu2018group}, and Instance Normalization \cite{DBLP:journals/corr/UlyanovVL16}; the recently proposed Switchable Normalization \cite{luo2018switch} meta-learns which normalization method to use during training. While these methods each have their merits, they do not resolve the sensitivity of DNNs to image-degradation because these have properties that are not observed by the network..

\begin{figure*}[h!t]
    \centering
    \includegraphics[width=0.9\textwidth]{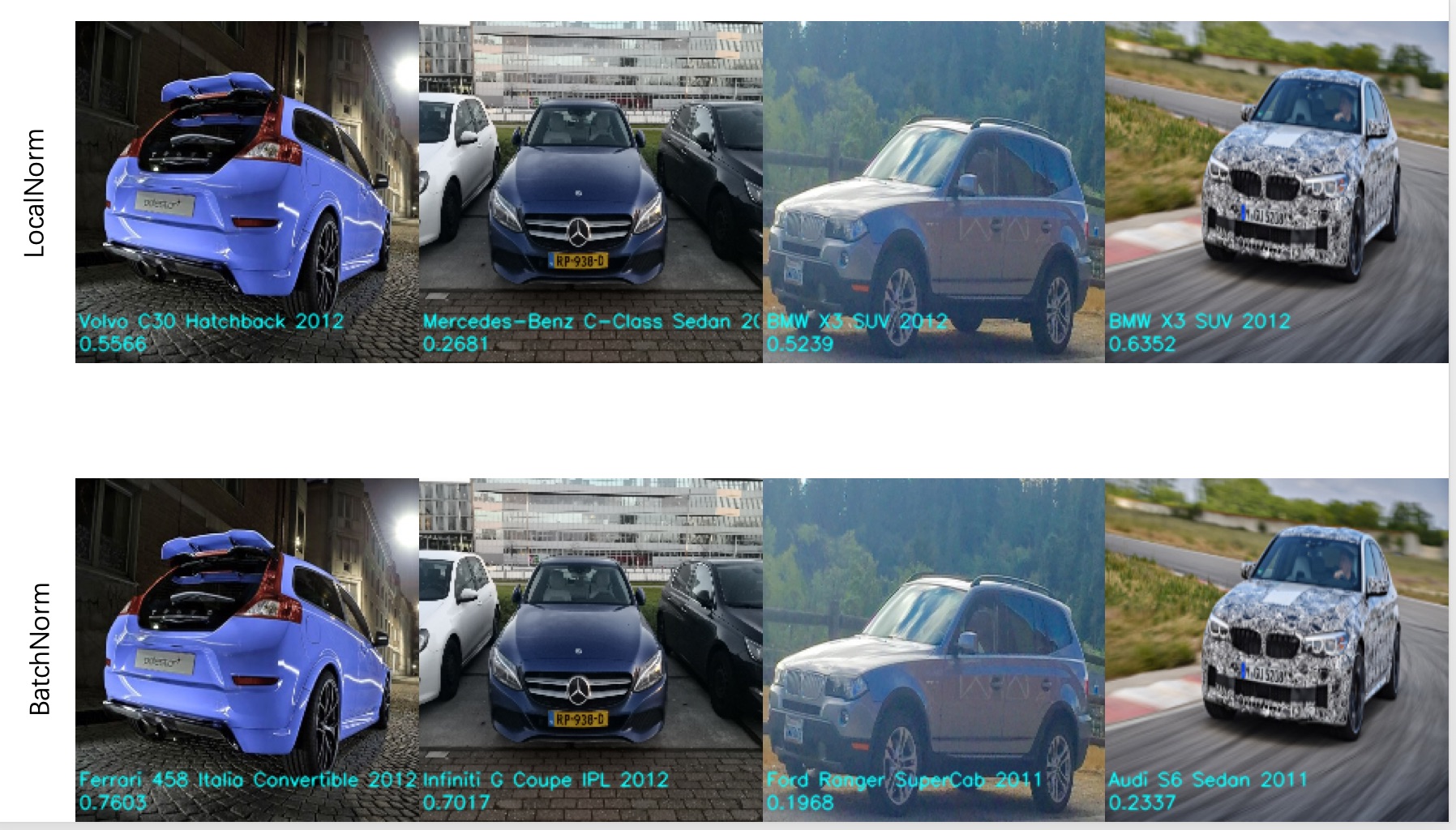}
    \caption{Examples of LocalNorm and BatchNorm classification on poorly lit or camouflaged images collected from the web.}
    \label{fig:demo}
\end{figure*}

Here, we propose a local variant of \BNp (BatchNorm), \LNp (LocalNorm), inspired by the continuous adaptation of spiking neurons to local temporal contrast \cite{mensi2016}: we observe that the mean and variance in channel activity changes when images are subjected to noise-related degradation. Figure \ref{fig:agn-rgb} shows an example of how the addition of Gaussian Noise flattens the color distribution for each channel in an image - other types of noise similarly affect the summary statistics, see Appendix. 
To increase the summary image statistical variance of the world from which the network learns, LocalNorm regularizes the normalization parameters during training by splitting the Batch into Groups, each with their own normalization scaling parameters. At test-time, the local channel statistics are then computed on the fly, either over a single image or a set (batch) of images in the test-set. 

We show that DNNs trained with LocalNorm normalization are much more robust to image degradation: the trained networks exhibit strong performance for unseen images with noise conditions that are not in the training set. 
An example is shown in Figure \ref{fig:demo}, where poorly lit or camouflaged images of cars are misclassified in the network using BatchNorm and correctly classified by the same network architecture using LocalNorm. We also find that networks drastically improves classification of distorted images in general, as measured on the CIFAR10-c dataset \cite{hendrycks2018benchmarking}, and we suggest a simple data augmentation scheme to improve summary statistics of small images. LocalNorm is straightforward to implement, also for networks already trained with standard BatchNorm - we show how a trained ResNet152 network trained further with LocalNorm improves accuracy for the Stanford Car dataset. Training networks from scratch, we show that LocalNorm achieves the same or slightly better performance as BatchNorm (and modern variants) on image classification benchmarks at little additional computational expense.  


\section{Related work}
Lighting and noise conditions can vary wildly over images, and various pre-processing steps are typically included in an image-processing pipeline to adjust color and reduce noise. In traditional computer vision, different filters and probabilistic models for image denoising are applied \cite{motwani2004survey}. Modern approaches for noise removal include deep neural networks, like Noise2Noise \cite{DBLP:journals/corr/abs-1803-04189}, DURR \cite{DBLP:journals/corr/abs-1805-07709}, and a denoising AutoEncoder \cite{vincent2010stacked} where the network is trained on a combination of noisy and original images to improve its performance on noisy dataset thus increasing the  networks' robustness to image noise and also to train a better classifier. However, as noted in \cite{geirhos2018generalisation}, training on images that include one type of noise in DNNs does not generalize to other types of noise. 


\begin{figure*}[h!t]
	\begin{multicols}{2}
    \centering
    \includegraphics[width=2\columnwidth]{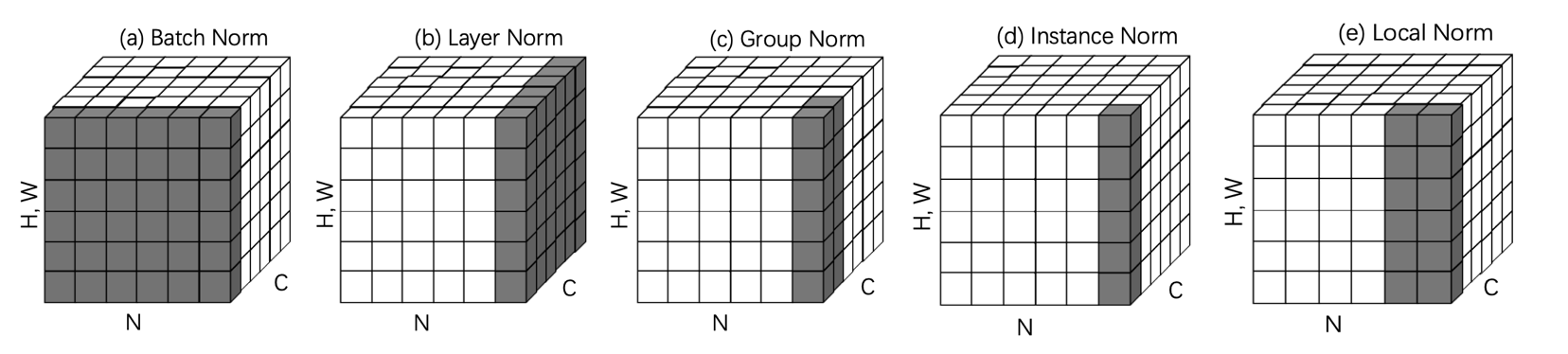}
    \end{multicols}
    \vspace*{-8mm}
    \caption{\textbf{Variants of Normalization Methods.} Each cube corresponds to a feature map tensor, with N as the batch axis, C as the channel axis, and (H, W) as the spatial axes -- height and width. The pixels in gray are normalized by the same mean and variance, computed by aggregating the values of these pixels.}
    \label{fig:Norms}
\end{figure*}

\subsection{Neural Normalizing techniques}
Normalization is typically used to rescale the dynamic range of an image. This idea has also been applied to deep learning in various guises, and notably Batch Normalization (\textbf{BatchNorm}) \cite{ioffe2015batch} was introduced to renormalize the mean and standard deviation of neural activations using an end-to-end trainable parametrization.

\textbf{Normalization techniques.}
A Normal-based normalization is generally computed as 
$$\hat{x_i} = \frac{x_i-\mu_i}{\sigma_i+\epsilon}*\gamma+\beta$$
where the $x_i$ is a part of feature tensor $X=\{\cup x_i\}$ computed by the previous layer and $\gamma$ and $\beta$ are the (trainable) scaling parameters. For normal 3-Dimensional image like RGB and GBR, $i=(i_N,i_W,i_H,i_C)$ is a 4D vector indexing the feature in $[N,W,H,C]$ order where $N$ is the batch size(number of images per batch), $H$ and $W$ are the spatial height and width axes, and $C$ is the channel axis. 

The space spanned by $N,H,W,C$ can be subdivided and subsequently normalised in multiple ways. We call the subdivision, the elements on which this normalization is performed, a group $Gk$: different forms of input normalisations can be described as dealing with different groups. The mean $\mu_k$ and standard deviation $\sigma_k$ of the certain computation group $G_k$ are computed as:
$$\mu_k = \frac{1}{m}\sum_{x_j\in G_k}x_j; \quad \sigma_k = \sqrt{\frac{1}{m}\sum_{x_j\in G_k}(x_j-\mu_i)^2+\epsilon}$$
where $\epsilon$ is a small constant like $10^{-7}$. The computation group $G_k$ (where $X=\{\cup G_k \mid k=1,2,\ldots K \}$)is a set of pixels which shares the mean $\mu_k$ and std $\sigma_k$, and $m$ is the size of the group $G_k$. BatchNorm and its variants can be mapped to a computational group along various axes (Figure \ref{fig:Norms}).

{\bf Batch Normalization (BatchNorm) }
was developed to ease training and improve convergence speed and generalization ability of deep neural networks. In \ref{fig:Norms}(a), for each channel, BatchNorm computes $\mu$ and $\sigma$ along the $(N,H,W)$ axes.
The computational group of BatchNorm comprises of all the pixels (inputs) from all $N$ batch samples sharing the same channel index. We can write this as $G_k = \{ p | p_c=i_c, c\in \{1,2,3,\ldots C\} \}$, where $p$ denotes the pixel and $p_c$ the pixel's channel index. 



{\bf Layer Normalization (LayerNorm)} \cite{ba2016layer} was designed to solve BatchNorm's dependence on the batch size, and as a smart way to apply a normalization method on recurrent networks. LayerNorm estimates the statistical features of one sample, which could also correspond to an input of a time step in sequence inputs (Figure \ref{fig:Norms}(b)). For each input sample, LayerNorm calculates ($\mu$ and $\sigma$) along the $(H,W,C)$ axes: as for BatchNorm, the computational group of LayerNorm can be defined as $G_k = \{ p | p_n=i_n, n\in \{1,2,3,\ldots N\} \}$. 

{\bf Group Normalization (GroupNorm)} \cite{wu2018group} was designed to enable the use of larger batches.
 In general, the use of larger batch sizes improves the generalization ability of the network and accelerates the training process \cite{DBLP:journals/corr/abs-1711-00489,DBLP:journals/corr/GoyalDGNWKTJH17}. 
 Large batch sizes however are typically limited by the locally available computational resources. 
 Group normalization computes summarizing statistics only over a subset of channels (the group; Figure\ref{fig:Norms}(c)), normalizing the computational group along the $(H,W,C/K)$ axes. The computational group for GroupNorm is thus defined  as $G_k = \{ p|p_n=i_n, \lfloor \frac{p_c}{C/K} \rfloor=\lfloor \frac{i_c}{C/K} \rfloor \}$. 
 
{\bf Instance Normalization (InstaNorm) }
\cite{DBLP:journals/corr/UlyanovVL16,DBLP:journals/corr/UlyanovVL17} was created for style transfer and quantity improvement. InstaNorm normalizes pixels of one sample in a single  channel (Figure\ref{fig:Norms}(d)). The InstaNorm computational group is defined as $G_k = \{p|p_n=i_n,p_c=i_c,n\in\{1,2,\ldots,N\},c\in\{1,2,\ldots,C\}\}$. 

{\bf Switchable Normalization (SwitchNorm)}  \cite{luo2018switch}  was proposed as the linear combination of BatchNorm, LayerNorm and InstaNorm:
 in the SwitchNorm layer, the relative weighing of each kind of normalization method is adjusted during the training process. This allows the network to learn the right type of normalization at the right place in the network to improve performance; this does come however at the expense of a sizable increase in parameters and computation.

\section{Local Normalization (LocalNorm)}



We develop LocalNorm to improve the robustness of DNNs to various noise conditions.
For BatchNorm, the  mean $\mu$ and std $\sigma$ are calculated along all training samples in a channel and then fixed for evaluation on test images; as noted however, when the (test) image distribution changes, these statistical parameters will drift. As a result, DNNs with BatchNorm layers are sensitive to input that deviates from the training distribution, including noisy images.

\begin{figure}[!hb]
\centering
\includegraphics[width=.85\columnwidth]{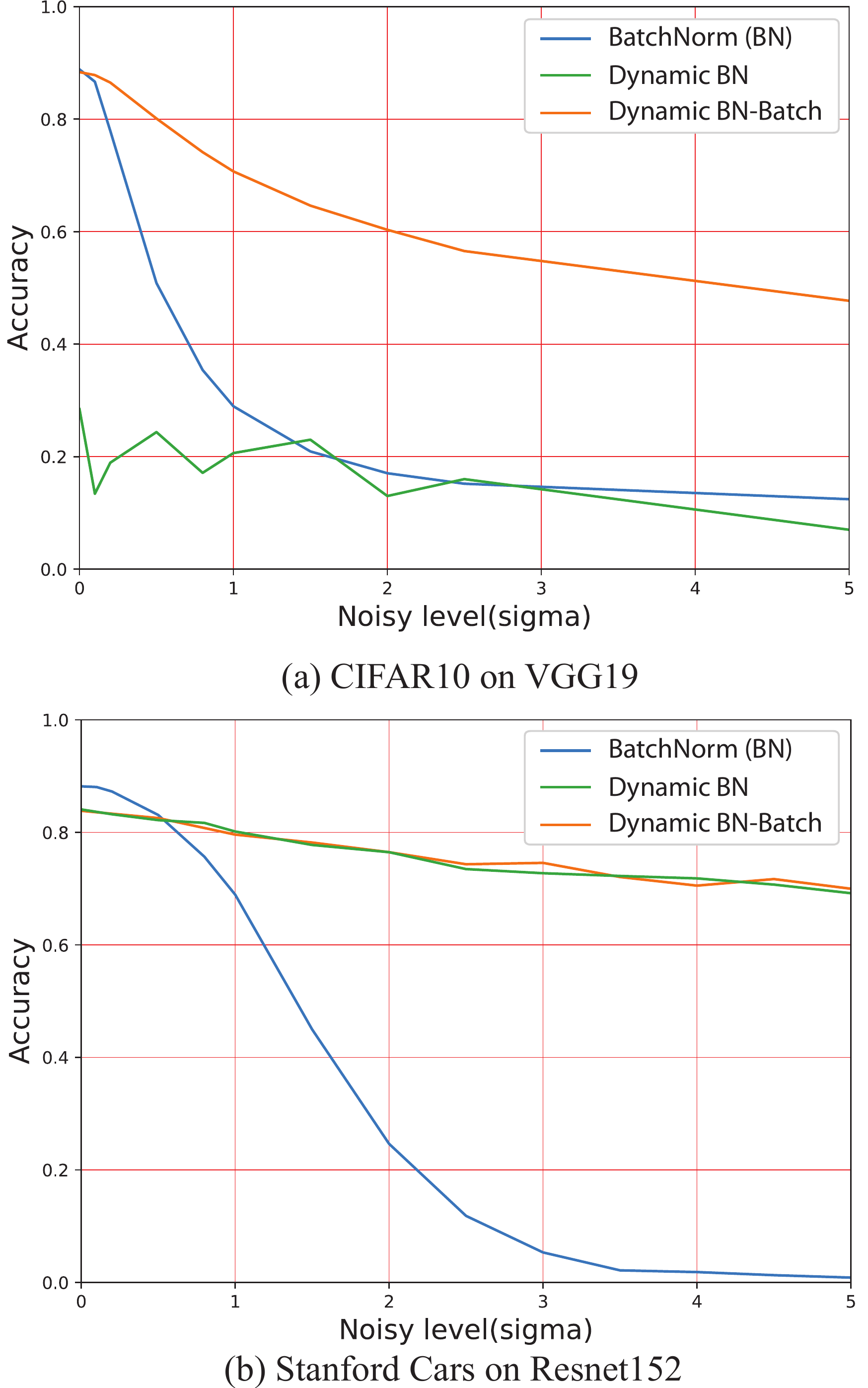}
    \caption{\textbf{AGN on CIFAR10 and Stanford Cars dataset.} Performance of a VGG19 network applied to Cifar10 (a) and a Resnet152 network to the Stanford Cars dataset (b) where the test-images are subjected to increasing amounts of image degradation, here in the form of Additive Gaussian Noise. Blue: accuracy for standard Batch Normalization. Orange: accuracy on dynamic Batch Normalization evaluated on single images. Green: accuracy on dynamic Batch Normalization with summary statistics computed over a batch of test-images.} 
    \label{fig:1groupBN}
\end{figure}

Simply computing the summary statistics on-the-fly, to account for a potential drift, only partly solves the problem:
in Figure \ref{fig:1groupBN}, we show what happens when the mean $\mu$ and std $\sigma$ are computed as dynamical quantities also at test time for standard benchmarks CIFAR10 and Stanford Cars, using modern deep neural networks (for details, see below). For each test image (or batch of test images) we compute ($\mu,\sigma$), for increasing noise (here for added Gaussian noise). For CIFAR10, Figure \ref{fig:1groupBN}a, we find that using single test images when evaluating gives poor results, as the small (32x32) images do not result in channel activity sufficient for effective summarizing statistics (Dynamic BN). However, computing these statistics over a batch shows a marked improvement (Dynamic BN-Batch): then, test accuracy exceeds standard BatchNorm for noisy images, at the expense of a slight decrease in accuracy for noiseless images. For the large images in Stanford Cars, we see that dynamically computing ($\mu,\sigma$) at test time even for single images drastically improves accuracy (Figure \ref{fig:1groupBN}b); the actual classification accuracy absent noise however drops. While computing summary statistics over a batch at test-time is feasible for benchmarking purposes, real world application would correspond to for example using a video stream, which would however substantially increase computational cost and latency.

\begin{figure*}[h!t]
    \centering
    \includegraphics[width=\textwidth]{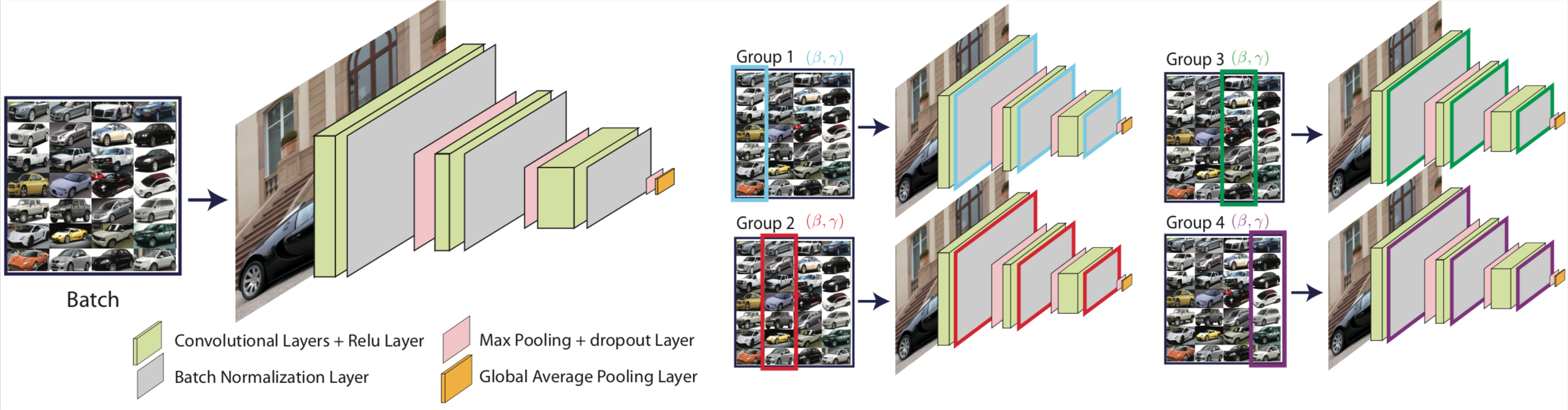}
    \caption{LocalNet. A deep network with standard batch normalization computes single summary statistics over the entire batch. In LocalNorm, summary statistics are computed over groups, where each group $k$ is associated with its own scaling parameters $\beta_k,\gamma_k$ (while sharing the all other network parameters), and summary statistics ($\mu_k,\sigma_k$) are dynamically computed also at test-time on the test-images.}
    \label{fig:localnet}
\end{figure*}

In {\bf LocalNorm}, we regularize the normalization layer for variations in $\mu$ and $\sigma$. The aim is to make the trained architecture less sensitive to changes in these statistics at test-time, such that we can dynamically recompute $\mu$ and $\sigma$ on test-images. We divide the Batch into separate Groups $G_k$ for which we each compute summarizing statistics $\mu_k,\sigma_k$ and associate separate scaling parameters $\gamma_k$ and $\beta_k$ with each Group (illustrated in Figure \ref{fig:localnet}). As shown in Figure\ref{fig:Norms}(e), for LocalNorm the computational group is defined along the $(N/K,H,W)$ axes: $$G_k = \left \{ p|p_c=i_c, \lfloor \frac{p_n}{N/K} \rfloor=\lfloor \frac{i_n}{N/K} \rfloor \right \}.$$ 
Effectively, each computational group can be regarded as a separate network sharing most parameters, where inputs are passed randomly through one such network during training. 

As noted, for BatchNorm the channel summary statistics $\mu,\sigma$ are taken as fixed from the training set after training. For LocalNorm, we recompute these statistics at test-time: this naturally incorporates changes in the image statistics, and the Group-induced regularized normalization ensures that the network also performs well for different such summary statistics.

Since LocalNorm provides both multiple independent Groups and computes summary statistics at test-time, there are different variants for classifying a novel image at test-time. Ideally, a single new image is passed through a randomly selected Group, such that summary statistics are computed on the fly only on this single image ($Single$). A second method is to do the same, but pass a single image through all Groups and then use voting to determine the classification ($Single-Voting$). A third method is to collect the number of images corresponding to the Group size ($Voting$), or use a set of images corresponding to the Batch size ($Batch$). For benchmark testing, $Batch$ is the fastest evaluation method, whereas $Voting$ is the computationally most desirable method for real-world application. 

\subsection{Implementation}
LocalNorm is easily implemented in auto-differentiation frameworks like \textbf{Keras} \cite{chollet2015keras} and \textbf{Tensorflow} \cite{abadi2016tensorflow} by adapting a standard batch normalization implementation\footnote{code available at \url{https://github.com/byin-cwi/LocalNorm1}}. For multi-GPUs, LocalNorm can map computational groups on separate GPUs which can accelerate training and allow the training of larger networks. In a variant of transfer learning \cite{pan2010survey}, it is straightforward to adapt a model pre-trained with BatchNorm by replacing all BatchNorm layers with LocalNorm layers initialized with the BatchNorm parameters, and then continue training.

\section{Image Noise}
We test LocalNorm in a Noisy-object classification task where synthetic Gaussian, Poisson and Bernoulli noise is added to images, as in Noise2Noise \cite{DBLP:journals/corr/abs-1803-04189}. All three kinds of independent noise $\xi$ are added on each channel of the image $x_c$ as follows:

For {\bf Additive Gaussian Noise (AGN)}, Gaussian noise with zero mean is added to the image on each channel, defined as $\hat{x_c} = x_c(1+\xi),\xi \sim Gaussian(0,\sigma_n)$.  

\begin{figure*}[!ht]
\centering
    \begin{multicols}{2}
	\includegraphics[width=1.0\textwidth,scale = 1.5]{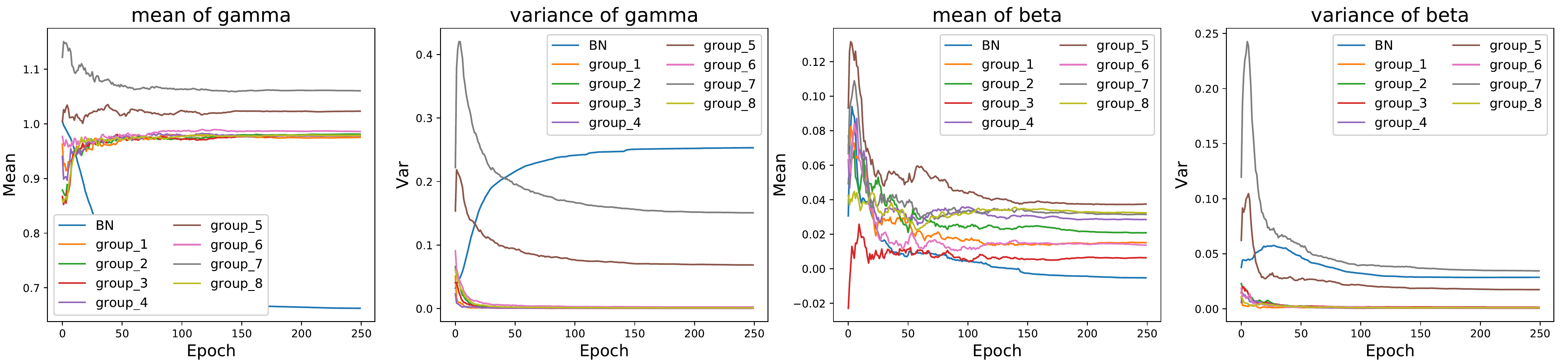}
    \end{multicols}
    \vspace*{-8mm}
    \caption{Development of mean and variance of the scaling parameters $\gamma$ and $\beta$ for LocalNorm Groups (group\textunderscore x) and BatchNorm (BN) during training on CIFAR10.}
    \label{fig:why}
\end{figure*}

{\bf Additive Poisson Noise (APN)} is one of the most dominating noise sources in photographs, and is easily visable in low-light images.
APN is a type of zero-mean noise and is hard to remove by pre-processing because it is distributed independently at each channel. Mathmatically, APN is computed as  $\hat{x_c} = x_c + 255\xi$ or $\hat{x_c} = x_c(1+\xi)$ $\xi \sim Poisson(0,\sigma_n)$, where $\sigma_n \in \lbrack0, 1\rbrack$.





{\bf Multiplicative Bernoulli Noise (MBN)} removes some random pixels from the image with probability $\sigma_n$. MBN defined by $\hat{x} = x\xi,\xi \sim Bernoulli(\sigma_n)$.

\section{Experimental Results}

\subsection{Benchmark Accuracy}
We apply LocalNorm to a number of classical benchmarks: MNIST \cite{lecun1998gradient}, CIFAR10 \cite{krizhevsky2009learning}, and Stanford Cars \cite{KrauseStarkDengFei-Fei_3DRR2013}, and compare with other normalization methods. Where useful, we evaluate the benchmarks using all four different types of LocalNorm evaluation methods; when not explicitly mentioned otherwise, the application of LocalNorm refers to the $Batch$ evaluation method. 


Results for all three normalization methods (BatchNorm, SwitchNorm and LocalNorm) are shown in Table \ref{table:Acc} using otherwise identical network architectures, where we evaluate LocalNorm with LocalNorm-Single, LocalNorm-Batch and LocalNorm-Voting. For BatchNorm, SwitchNorm, LocalNorm-Batch and LocalNorm-Voting, we achieve near state-of-the-art accuracy on the original datasets, where in 3 our of 4 cases, LocalNorm-Voting and LocalNorm-Batch outperform BatchNorm and SwitchNorm. The improvement for CIFAR10 using the VGG architecture with LocalNorm-Voting in particular stands out, as accuracy improves from 88.8\% to 95.3\%; no such improvement is observed for the ResnNet32 architecture, and only a slight improvement for the ResNet152 as applied ot Stanford Cars. We also observe that for the small images in CIFAR10, evaluating test-images using only a single image at a time (LocalNorm-Single) gives poor results. Comparing training time, for CIFAR10, we find that LocalNorm incurs only a small computational cost (10-20\%), while SwitchNorm proves much more computationally expensive (Table \ref{table:Acc}). 


For MNIST, we designed a standard DNN (Input-16c-16c-32c-32c-512d-1024d-output), we set the batch size to 100; for LocalNorm, we divide the batch into 10 computational groups with 10 images each group.
For CIFAR10, we use two classical network architectures -- VGG19 and ResNet32. The classical {\bf VGG19} network architecture  \cite{DBLP:journals/corr/SimonyanZ14a} is often used as a baseline to test new network architectures. 
{\bf Residual Networks}, or {\bf ResNets} \cite{he2016deep} have achieved state-of-the-art accuracy on many machine learning datasets, and ResNet32 (a ResNet with 32 Layers) achieves competitive results on the CIFAR10 dataset \cite{zhang2018three}. We use a batch size of 128, as in most recent state-of-the-art models. For LocalNorm, we divide the batch into 8 computational groups with 16 images per group by default.

\begin{table}[!ht]
\resizebox{\columnwidth}{!}{%
\begin{tabular}{|c|c|c|c|c|}
\hline
\textbf{} & \textbf{MNIST} & \textbf{CIFAR10-VGG} & \textbf{CIFAR10-ResNet} & \textbf{Stanford-Car} \\ \hline
\textbf{BatchNorm} & $99.60\%$          & $88.83\%$          & $91.74\%$          & $88.17\%$ \\ \hline
\textbf{SwitchNorm} & $99.53\%$         & $57.39\%$          & \textbf{91.88}$\%$ & $87.34\%$ \\ \hline
\textbf{LocalNorm-Single} &   $98.46\%$     & $65.88\%$          & $32.33\%$          & $88.39\%$ \\ \hline
\textbf{LocalNorm-Batch} & \textbf{99.67}$\%$ & 92.07$\%$ & $91.15\%$          & 89.34$\%$ \\ \hline
\textbf{LocalNorm-Voting} & $99.66\%$ & \textbf{95.29}$\%$ & $91.65\%$          & \textbf{89.58}$\%$ \\ \hline
\end{tabular}
}
\vspace*{-3mm}
\caption{The accuracy on original test dataset of each network with various types of normalization on each dataset and for different LocalNorm evaluation methods.}
\label{table:Acc}
\end{table}

\begin{table}[!hbt]
\resizebox{\columnwidth}{!}{%
\begin{tabular}{|c|c|c|c|c|}
\hline
\multirow{2}{*}{Models} & \multicolumn{2}{c|}{\textbf{VGG19}} & \multicolumn{2}{c|}{\textbf{ResNet32}} \\ \cline{2-5} 
                  & \textbf{Speed (s/epoch)} & \textbf{Paras} & \textbf{Speed (s/epoch)}      & \textbf{Paras}     \\ \hline
\textbf{BatchNorm}& 20s & 15,001,418 & 21s      & 470,218     \\ \hline
\textbf{LocalNorm}& 23s & 15,115,690 & 26s      & 473,610     \\ \hline
\textbf{SwitchNorm}& 30s & 15,001,496 & 56s     & 470,414     \\ \hline
\end{tabular}
}
\vspace*{-3mm}
\caption{Training speed and model size on VGG19 and ResNet32 with various Norms on CIFAR10 on an Nvidia Titan Xp.}
\label{table:time}
\end{table}

The Stanford Cars dataset contains 16,185 images of 196 classes of cars, and each image is large, similar to images in the ImageNet dataset, allowing us to compare LocalNorm to the other normalization methods when applied to large networks and large images. The training and test dataset are similarly large, and the images are taken under various conditions. We use ResNet152 for this dataset for improved accuracy; 16 images are trained as a batch and are divided into 4 groups for LocalNorm. For ResNet152, we use the pre-trained ImageNet weights from github\footnote{\url{https://gist.github.com/flyyufelix/7e2eafb149f72f4d38dd661882c554a6}} and then continue training this network with BatchNorm, SwitchNorm or LocalNorm. 




In Figure \ref{fig:why} we plot the development of mean and variance of the normalization scaling parameters $\gamma$ and $\beta$ for LocalNorm and BatchNorm (averaged over all channels) when training VGG19 on CIFAR10 using 8 Groups for LocalNorm. We see that LocalNorm converges to a spread of $\gamma$ and $\beta$ values during training. 

\begin{figure}[!th]
    \centering
    \includegraphics[width=0.85\columnwidth,height=8cm]{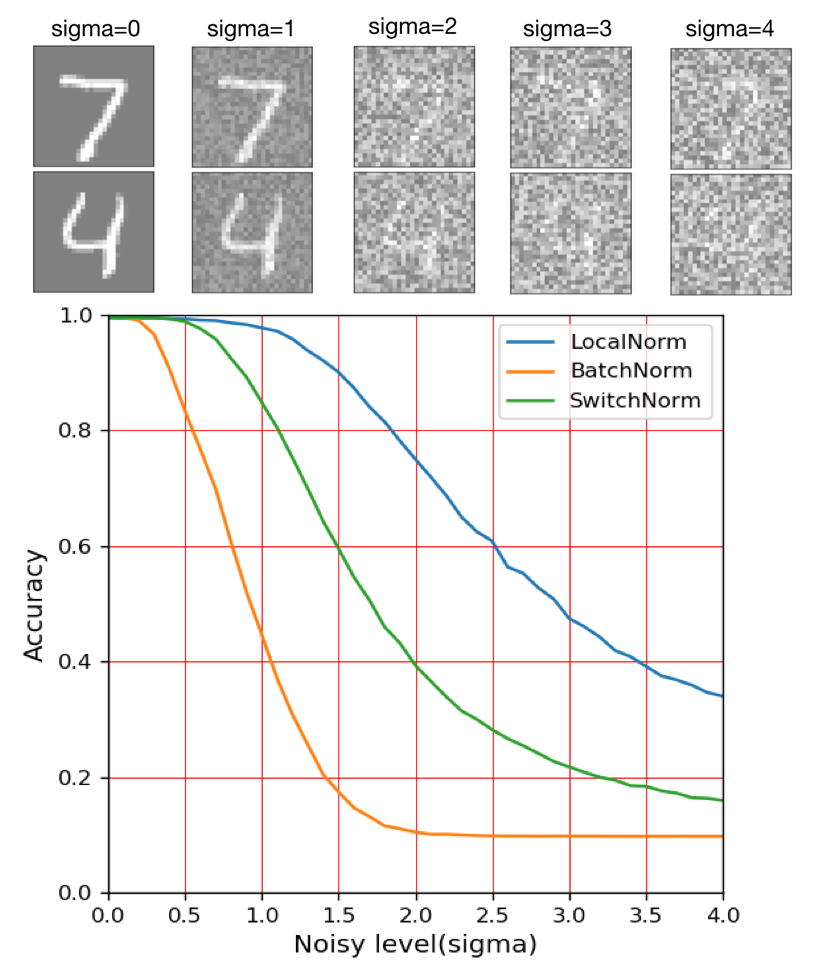}
    \vspace*{-3mm}
    \caption{\textbf{MNIST accuracy and noisy image.} The top row shows the image quality under different AGN, and Line graph plots the accuracy obtained for noise-degraded digits. LocalNorm-Batch was used for evaluation here.}
    \label{fig:MNIST}
\end{figure}

\subsection{Noisy Image degradation}
To measure noise robustness and noise generalization, we use the networks trained with various normalization methods and the original training dataset, and test them on images degraded with different levels of noise.

We evaluated the CIFAR10 and Stanford Cars dataset for all variants of LocalNorm, both where a batch of images is used at test-time ($Batch$ and $Voting$) to obtain summary statistics, and where only a single image at a time is used at test-time to obtain summary statistics ($Single$ and $Single-Voting$).

\begin{figure*}[!ht]
     \centering  
        \includegraphics[width=\textwidth]{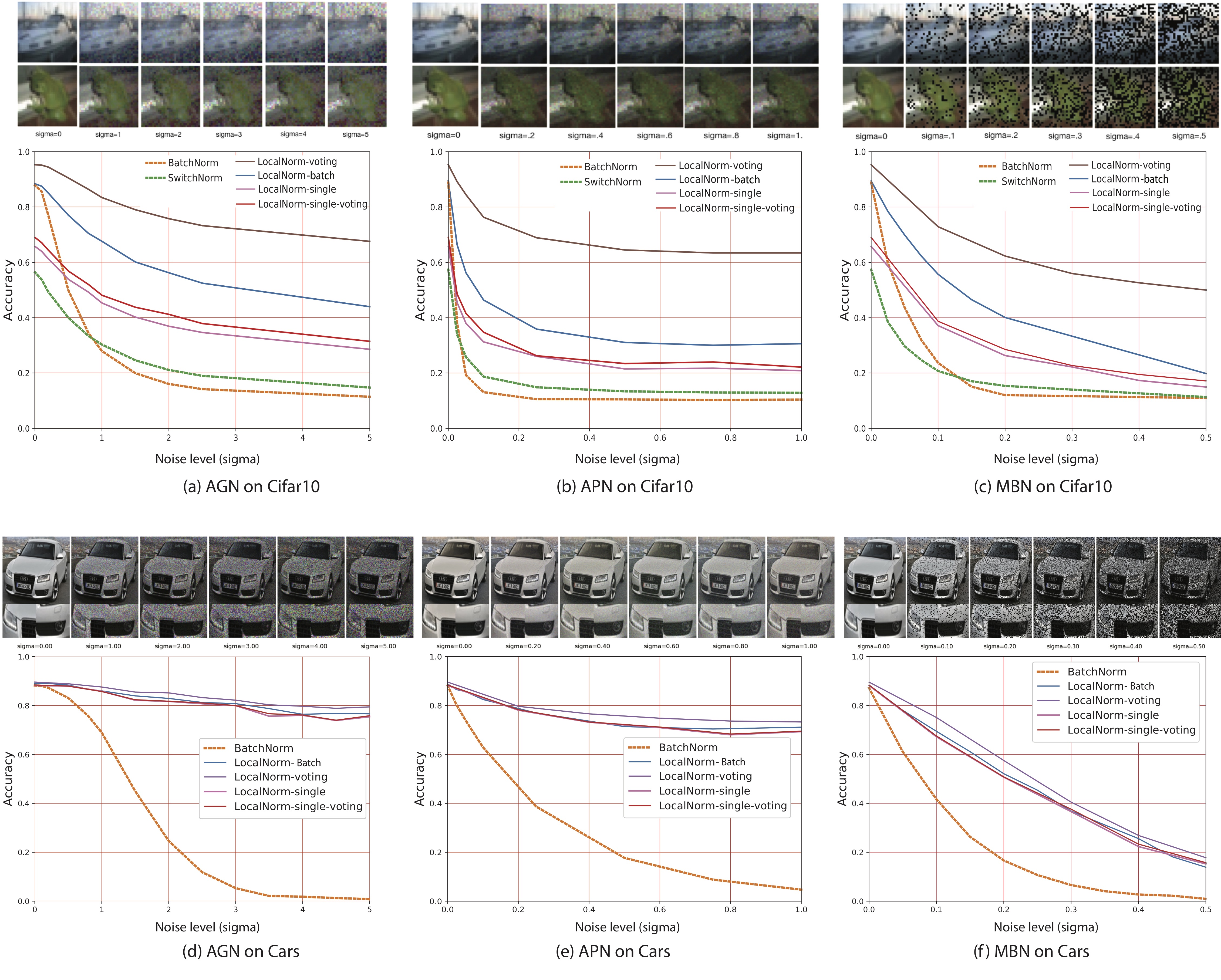}
        
\vspace{-0.1cm}
\caption{\textbf{Noise effect on CIFAR10 (a-c) and Stanford Cars datasets (d-f).} (a-c) Top row illustrates noise-degraded CIFAR10 images for different amounts of AGN, AGN and MBN respectively. Bottom row, line graphs plot corresponding network accuracy on degraded CIFAR10 images using a VGG19 network architecture; (d-f) same noise-degradations applied to the Stanford Cars images using a ResNet152 network architecture. Dashed line: BatchNorm (orange) and SwitchNorm}
\label{fig:noiseCifar10vgg}
\end{figure*}

\paragraph{MNIST}
In the MNIST dataset, images only have one channel. We apply AGN to MNIST to demonstrate DNN performance facing out-of-sample noise-degraded images. In Figure \ref{fig:MNIST}, we see that for all normalization methods, performance decreases when images become more degraded, e.g., for $\sigma_n=1$, the digit is clearly visible as is some noise. The performance of BatchNorm and SwitchNorm however decreases to $44.7\%$ and $84.9\%$ respectively, while LocalNorm still achieved an accuracy of \textbf{97.8}$\%$; for $\sigma_n=2$, where BatchNorm already yields random choice performance (around $10\%$), LocalNorm still performs with moderately reduced accuracy of $75.0\%$ (SwitchNorm obtains $39.3\%$). For very high noise levels, also difficult for humans, LocalNorm still outperforms SwitchNorm by a factor of two.


\paragraph{CIFAR10}
We tested VGG19 trained on CIFAR10 with various normalization methods on noisy test images degraded with AGN. Figure \ref{fig:noiseCifar10vgg}a shows that the accuracy when using BatchNorm decreases rapidly, achieving only $29\%$ accuracy for sigma=1. For the different types LocalNorm evaluation, we find that LocalNorm-Batch and LocalNorm-Voting substantially improve over BatchNorm and SwitchNorm, where for LocalNorm-Voting the network accuracy is $83\%$ at sigma=1, almost three times better than the BatchNorm-based network. Evaluation using only single images, LocalNorm-Single and LocalNorm-Single-Voting, while being more robust to noise, clearly underperform for noiseless data.
Similar observations apply for the other types of noise. For APN, both BatchNorm and LocalNorm's accuracy curve dropped sharply, while the LocalNorm still substantially outperforms BatchNorm and SwitchNorm in general (Figure \ref{fig:noiseCifar10vgg}b). For MBN in Figure \ref{fig:noiseCifar10vgg}c, both SwitchNorm and BatchNorm's accuracy drops exponentially and converge to random choice, while LocalNorm's performance decreases slower. 
We see the same performance order for a ResNet32 network applied to CIFAR10 (see Appendix, Figure \ref{fig:resnet32}).


\begin{figure*}[!h]
    \centering
    \includegraphics[width=0.9\textwidth,angle=0,scale=1.0]{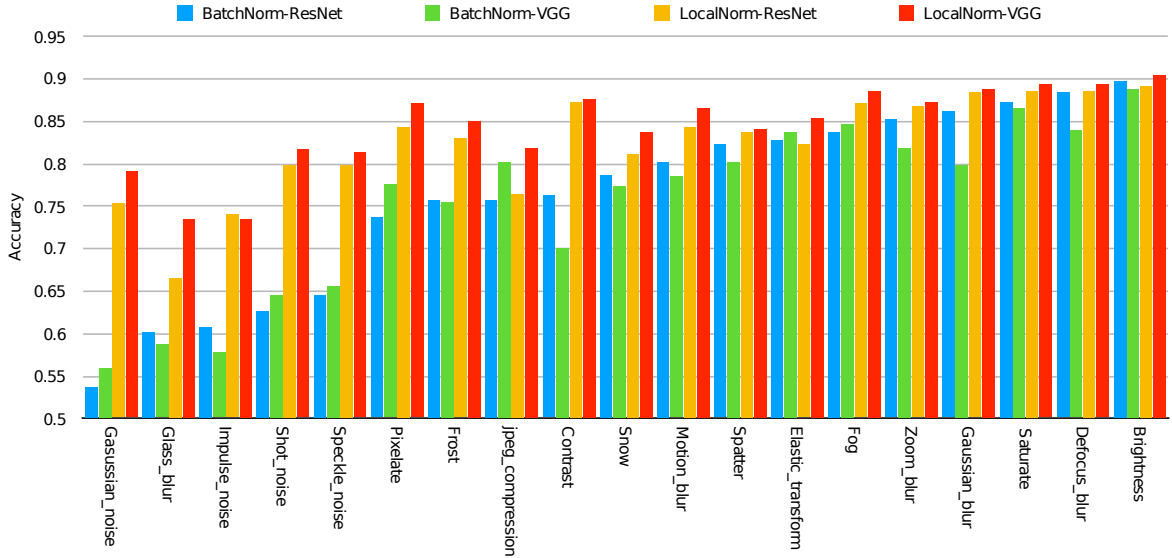}
    \caption{Comparison of LocalNorm-Batch to BatchNorm on the Cifar10-C dataset, for both Resnet32 and VGG19 networks, for all the different image corruption categories.}
    \label{fig:more_oise}
\end{figure*}

\paragraph{CIFAR10-c}
The Cifar10-C dataset was published specifically to test network robustness to image corruption \cite{hendrycks2018benchmarking}. It contains 19 types of algorithmically generated corruptions from noise, blur, weather, and digital categories. To evaluate robustness, the networks are trained on the original CIFAR10 dataset, and evaluated on the corrupted dataset using LocalNorm-Batch. The result are shown in Figure \ref{fig:more_oise}: we find that LocalNorm-Batch outperforms standard BatchNorm everywhere, with the largest improvements observed for those image corruptions that incur the largest performance drop (Noise, Blur).  We also see that LocalNorm improves the accuracy of the VGG-19 network much more than for the ResNet32 network, to the point that VGG becomes substantially more accurate than ResNet32.

\paragraph{Stanford Car Dataset}

    
For the large images in the Stanford Cars dataset, we find that when testing on noisy images (Figure \ref{fig:noiseCifar10vgg}d), all LocalNorm variants perform very similar, demonstrating that here, a single large image is sufficient to dynamically compute the summary statistics at test-time. LocalNorm maintains a test accuracy over $74\%$ under any tested level of AGN, while under BatchNorm accuracy declines sharply to $<20\%$ for sigma $>$ 2.5; a similar behavior is observed for APN (Figure \ref{fig:noiseCifar10vgg}e). For MBN, Figure \ref{fig:noiseCifar10vgg}f, the BatchNorm accuracy decreases exponentially while LocalNorm's performance declines essentially linearly\footnote{For Stanford Cars, we omitted data for SN as we obtained near-zero performance on noise-degraded images with the publicly available code.}. 

To directly investigate generalization ability under different noise levels, we computed the confusion matrix for each model under various conditions: this is shown in Figures \ref{fig:CM_AGN}-\ref{fig:CM_MBN} in the Appendix. 
In general, we find that networks using BatchNorm increasingly default classification to a select few classes for increasing noise levels, whereas for networks using LocalNorm this is not the case - classification becomes essentially random.  

	
	

%

\begin{figure*}
\centering
\begin{multicols}{2}
    \centering
    \includegraphics[width=2.\columnwidth]{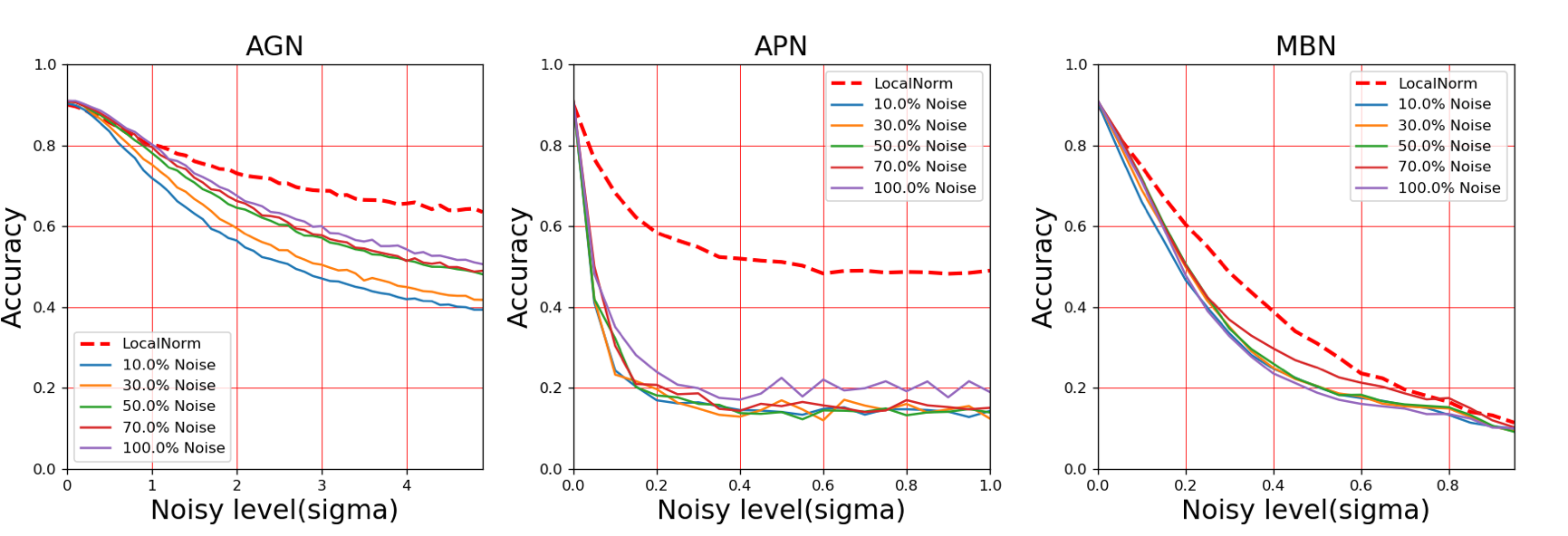}
\end{multicols}
\vspace*{-5mm}
    \caption{Training ResNet32 on a training set augmented with an increasing number of AGN images in CIFAR10, for sigma=1. 
    }
    \label{fig:addnoise}
\end{figure*}

\subsection{Single Image Data augmentation at test-time}
\begin{figure}[!th]
    \centering
    \includegraphics[width=0.85\columnwidth]{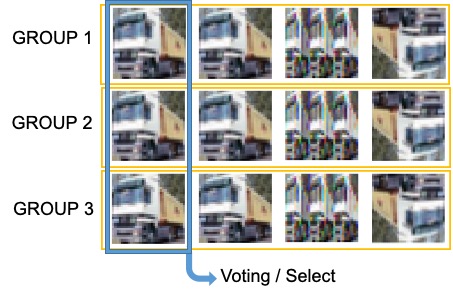}
    \vspace*{-3mm}
    \caption{The classification workflow of single image using data augmentation of the summary statistics through group expansion with rotated images. For a single image in each group, rotated versions are created and added to the group. Summary statistics are computed for the whole group, while for classification only single image is used (either from a randomly selected group --LocalNorm-Single-- or using voting -- LocalNorm-Singe-Voting.}
    \label{fig:rotate}
\end{figure}

To improve the performance of LocalNorm-Single and LocalNorm-Single-Voting evaluation on small images, a simple suggestion is to enrich the summary statistics. Here, we augment the data by adding rotated versions of the image to the computation group to enrich the summary statistics. We find that this trick drastically improves LocalNorm-Single and LocalNorm-Single-Voting for the small images of CIFAR10. For Cifar10 dataset, adding the rotated the image along the axis W and C could improve the single image performance, as shown in Fig\ref{fig:rotate}, this increases the details of the mean of the computational group.

\begin{figure}[!th]
    \centering
    \includegraphics[width=0.95\columnwidth]{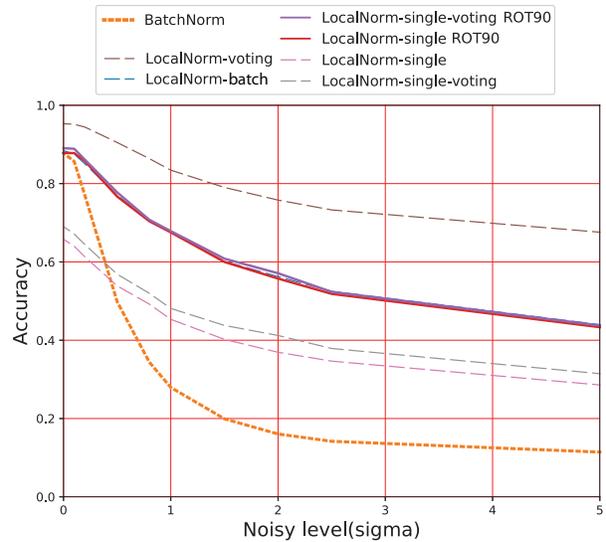}
    \vspace*{-3mm}
    \caption{Results of rotation-enhanced summary statistics. The `ROT90' lines indicate data-augmented evaluation.}
    \label{fig:rot90cifar10}
\end{figure}

During classification, the prediction is made for the original image, and rotated images are only used to compute the summary statistics. In Figure \ref{fig:rotate}; as before, this type of classification can be done by either voting the prediction of each group or selecting a prediction randomly as the final result. As show in Figure \ref{fig:rot90cifar10} for AGN, we find that for CIFAR10, thus enhancing the summary statistics for single image evaluation improves robustness and noiseless accuracy to the same level as LocalNorm-Batch - we observe the same for image degradation with APN and MBN (not shown).

While performance improves and such rotation allows a network to apply LocalNorm also to the small images of CIFAR10, this comes at the cost of filling one group or multiple groups with rotated images and computing the corresponding network activity. While this is a substantial increase in computational cost, there is no cost to training, and evaluation on such small images tends to be fast.



\subsection{Training effects}

{\bf Training on augmented noisy datasets. }
We next examine how network robustness improves when noisy AGN images are added to the {\em training} dataset. As can be seen in Figure \ref{fig:addnoise}, when testing on images with AGN or MBN noise, adding AGN noise samples in the training set does improve accuracy for BatchNorm-trained networks on noisy test-images. This AGN-noise network however hardly improves accuracy on test-data containing Poisson noise (APN) or Bernouilli noise (MBN), confirming the observation in \cite{geirhos2018generalisation} that noise is hard to generalize. Moreover, networks trained using LocalNorm without added noise samples still perform better, and we also find that for the noise-augmented BatchNorm network the test accuracy on the original dataset  is slightly reduced. In practice, it is next to impossible to cover all noise conditions in the training dataset, and training with many such added examples is computationally expensive.



\begin{figure*}[!hbt]
\centering
\begin{multicols}{2}
\centering
    \includegraphics[width=2.\columnwidth]{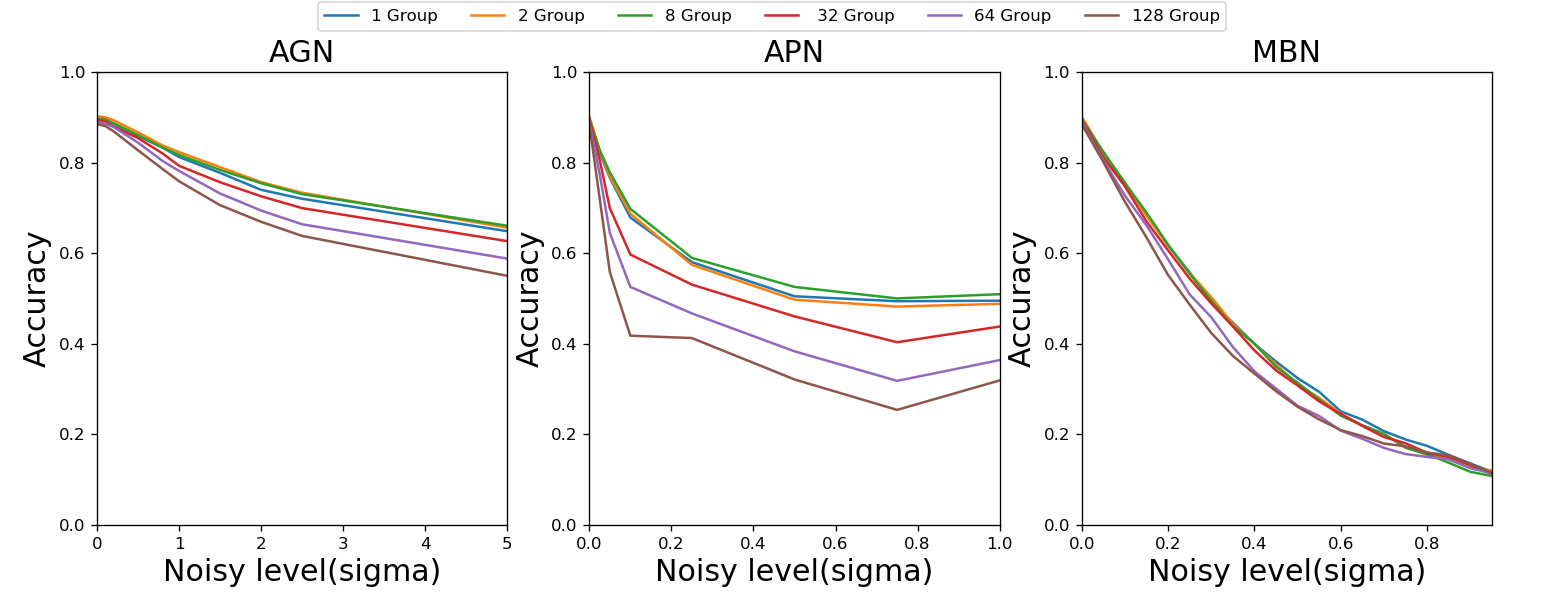}
\end{multicols}
\vspace*{-7mm}
    \caption{Group number's influence on LocalNorm-Batch performance on Cifar10 (using VGG19)}
    \label{fig:Group_sizeAPP}
\end{figure*}

{\bf Group size. }
LocalNorm has as a parameter the number of groups which, for a given batch size, determines the number of images in each group. While we did not extensively optimize for group number, we found that a small-ish number of groups, 4-8, performed best in practice for the batch sizes used in this study (Figure \ref{fig:Group_sizeAPP}).

\section{Conclusion}

We develop an effective and robust normalization layer--LocalNorm. LocalNorm regularizes the Normaliation layer during training, and includes a dynamic computation of the Normalization layer's summary statistics during test-time. The key insight here is that out-of-sample conditions, like noise degradation, will shift the summary statistics of an image, and the LocalNorm approach makes a DNN more robust to such shifts. 


We demonstrate the effectiveness of the approach on classical benchmarks, including both small and large images, and find that LocalNorm decisively outperforms both classical Batch Normalization and modern variants like SwitchNorm. We show that computing LocalNorm only has a limited computational cost with respect to training time, of order 10-20\%. LocalNorm furthermore can be evaluated on batches of test-images, and, for large enough images, also on single images passed through only a single group, then incurring the same evaluation cost as Batch Normalization. 
To enable the evaluation of small images one-at-a-time, we demonstrated the use or image rotation as a form of data augmentation to sufficiently improve the summary statistics. 
For more general type of image distortions, we find that using LocalNorm also makes networks substantially more robust, as evidenced by the results on the CIFAR10-c dataset.

\section*{Acknowledgments} BY is funded by the NWO-TTW Programme ``Efficient Deep Learning'' (EDL); the Titan Xp used for this research was donated by the NVIDIA Corporation.

\bibliography{yin_references}
\bibliographystyle{icml2019}

\appendix

\begin{figure*}[!hbt]
    \centering
    \includegraphics[width=\textwidth,height=3.6cm]{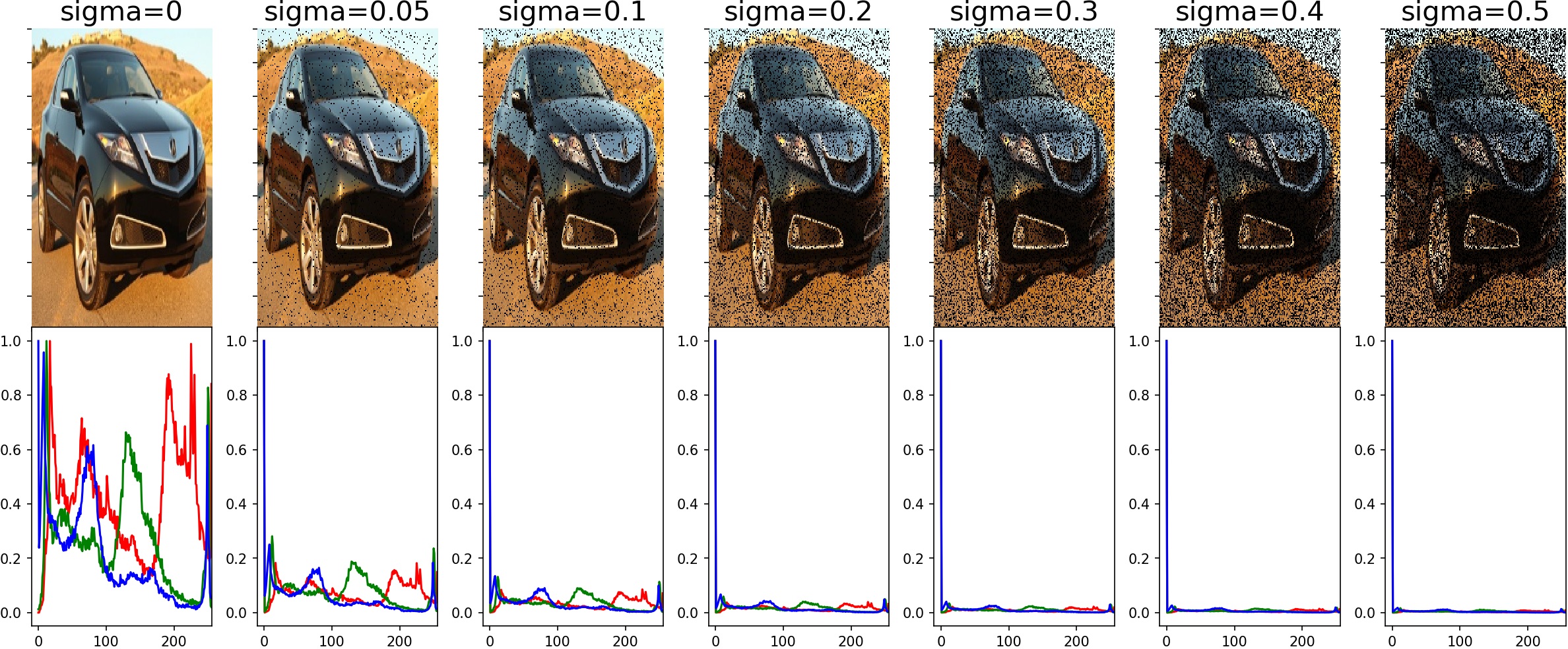}
    \caption{RGB-Histogram for MBN-type noise.}
    \label{fig:mbn-rgb}
\end{figure*}

\begin{figure*}[!hbt]
    \centering
    \includegraphics[width=\textwidth,height=3.6cm]{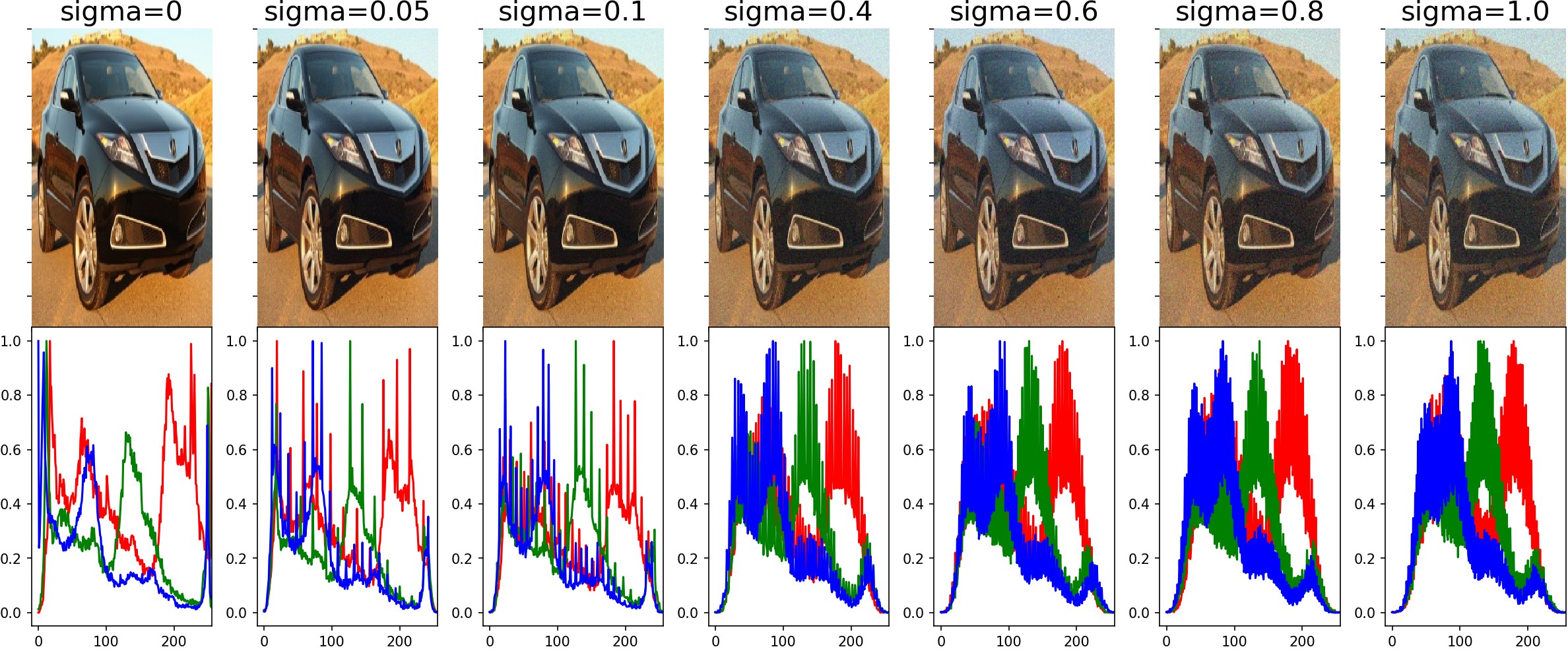}
    \caption{RGB-Histogram for APN-type noise.}
    \label{fig:apn-rgb}
\end{figure*}

\begin{figure*}[!hbt]
        \centering
        \includegraphics[width=0.98\textwidth]{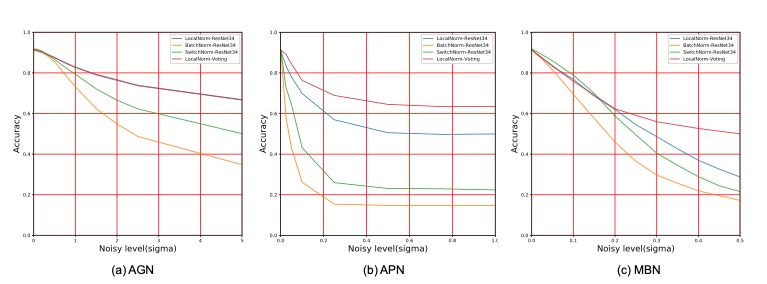}
    \caption{ResNet32 on CIFAR10. Evaluated for LocalNorm-Batch and LocalNorm-Voting.}
    \label{fig:resnet32}
\end{figure*}





\begin{figure*}[!hbt]
    \centering
    \includegraphics[width=\textwidth,angle=270,scale=1.2]{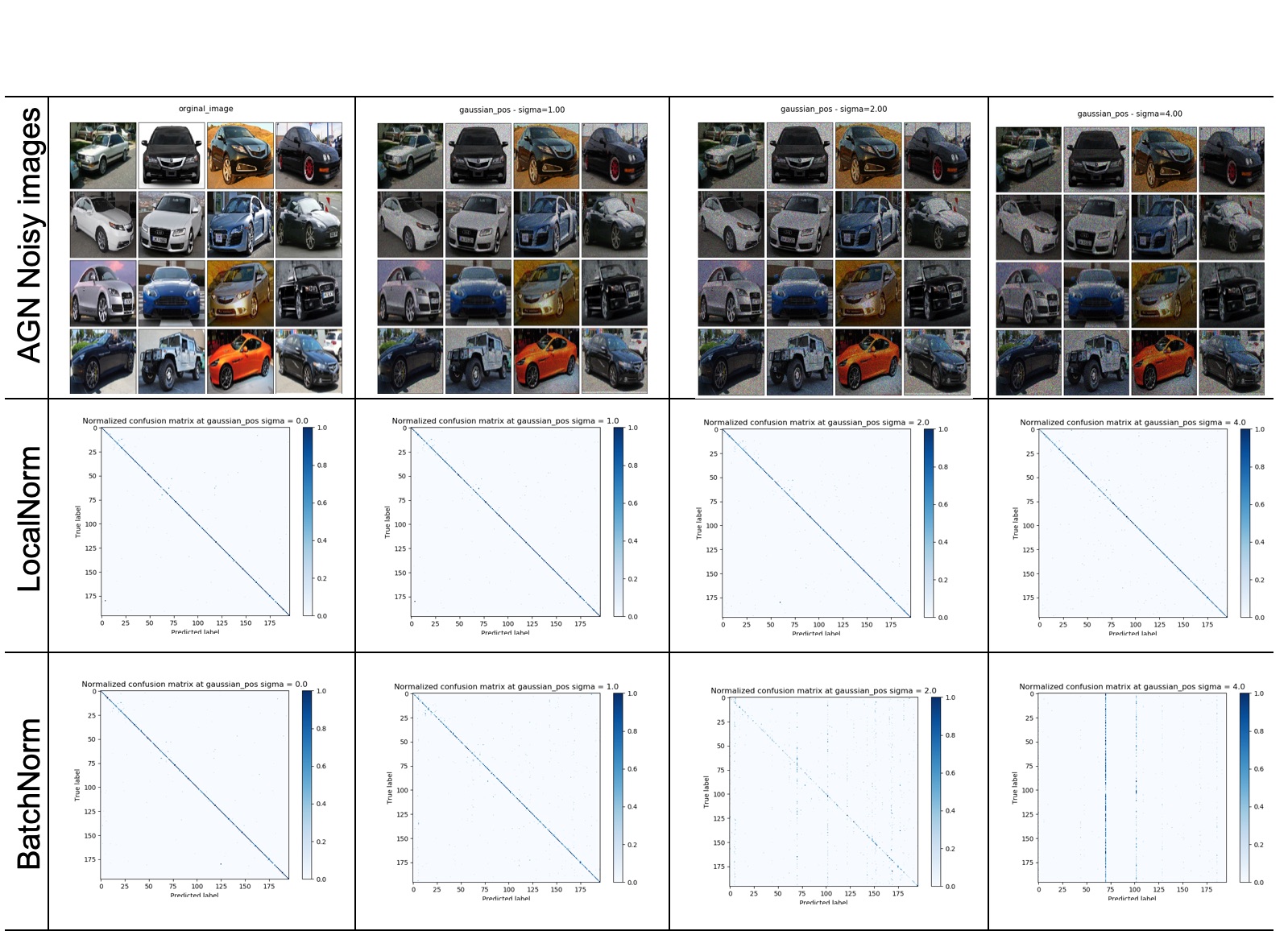}
    \caption{Confusion matrix under AGN noise for $\sigma = \{0,1.,2.,4.\}$}
    \label{fig:CM_AGN}
\end{figure*}

\begin{figure*}[!h]
    \centering
    \includegraphics[width=0.8\textwidth,angle=0,scale=1.0]{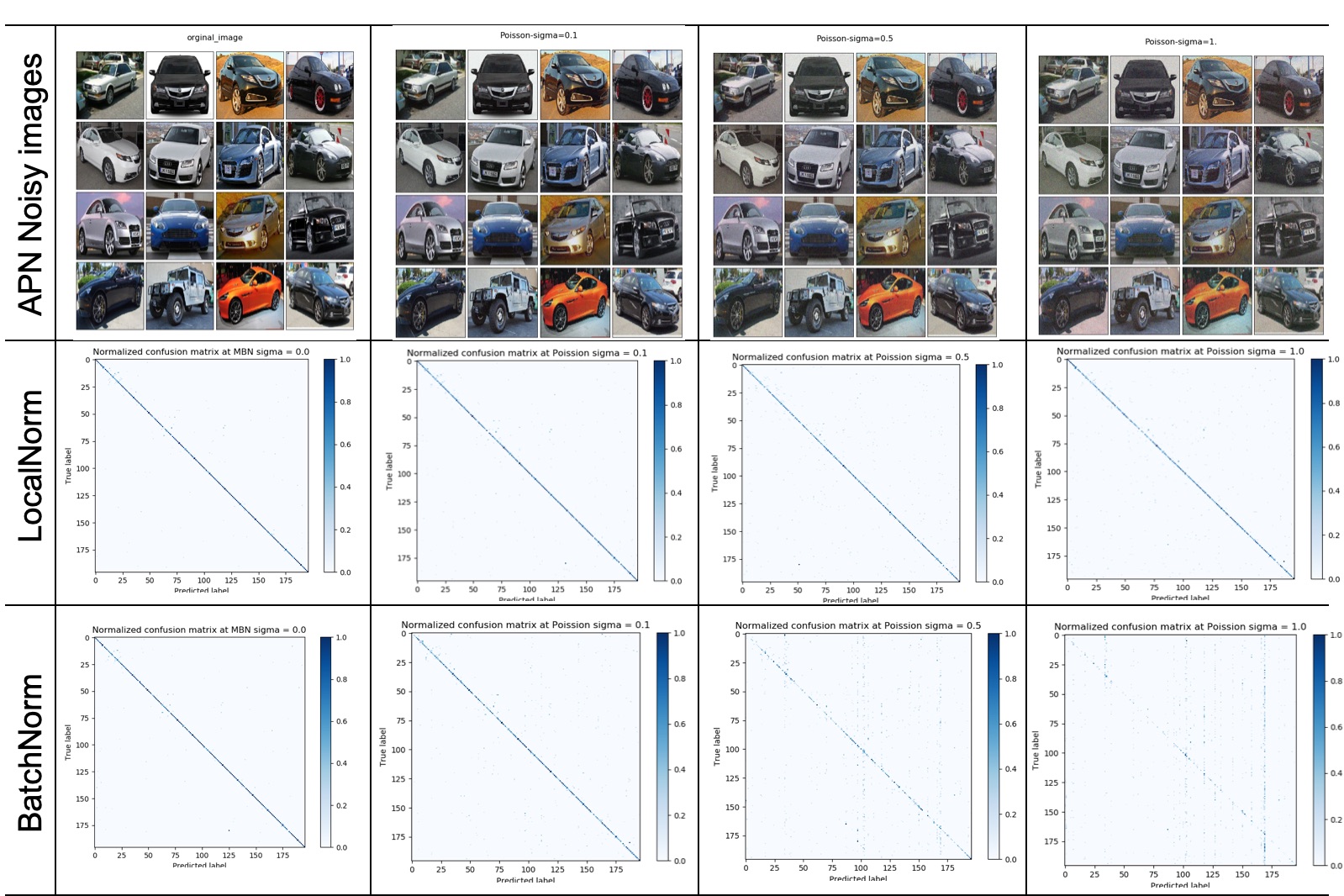}
    \caption{Confusion matrix under APN noise where $\sigma = \{0,.1,.5,1.\}$}
    \label{fig:CM_APN}
\end{figure*}

\begin{figure*}[!h]
    \centering
    \includegraphics[width=0.8\textwidth,angle=0,scale=1.0]{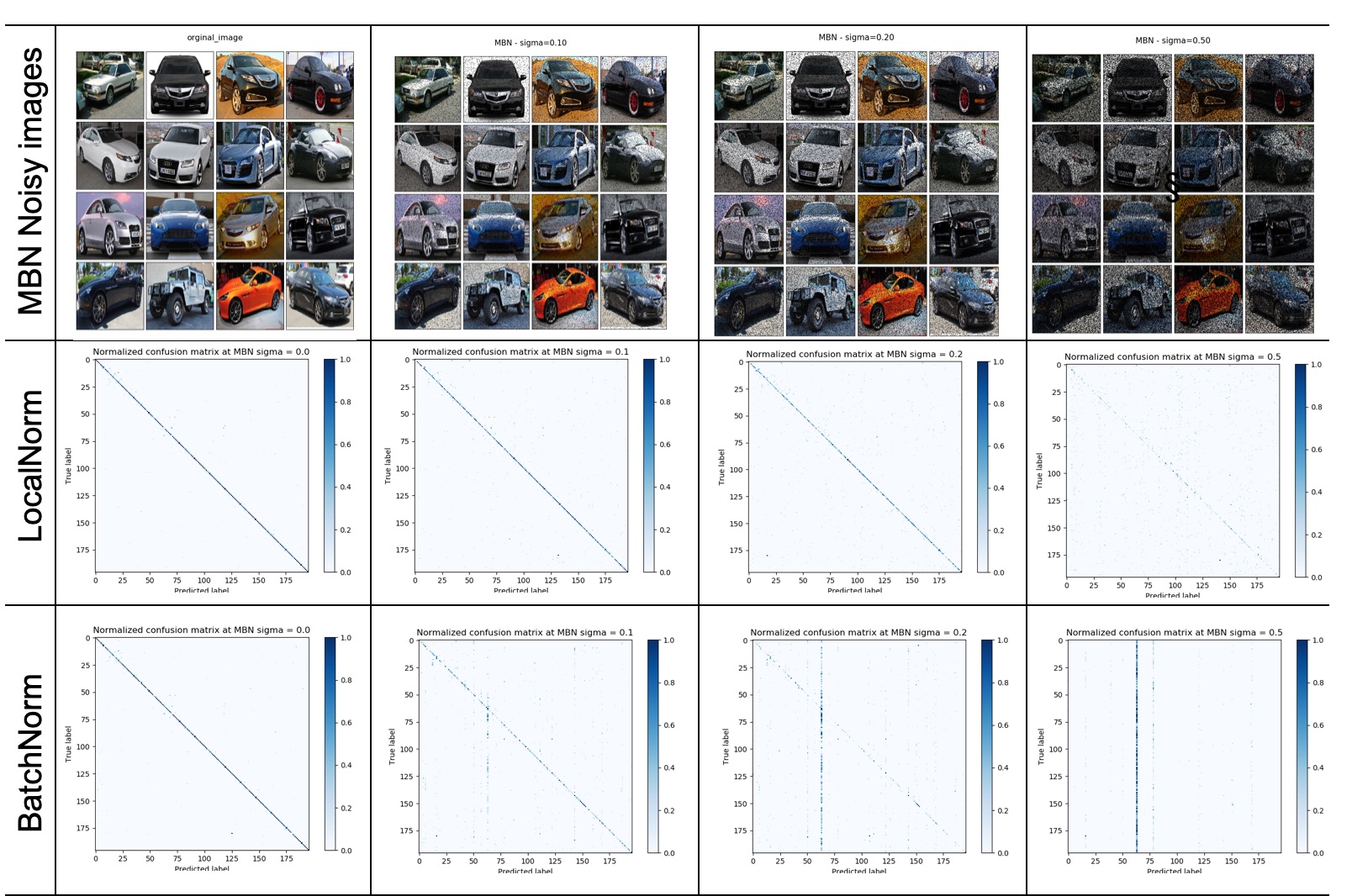}
    \caption{Confusion matrix under MBN noise where $\sigma = \{0,.1,.2,.5\}$}
    \label{fig:CM_MBN}
\end{figure*}


\end{document}